\documentclass[runningheads]{llncs}
\usepackage[utf8]{inputenc}
\usepackage{amsmath}
\usepackage{amsfonts}
\usepackage{algpseudocode}
\usepackage{graphicx}
\usepackage{url}
\usepackage[ruled]{algorithm2e}

\title{Optimizing Integrated Information with a Prior Guided Random Search Algorithm}
\titlerunning{Optimizing Integrated Information with a Random Search Algorithm}
\author{Eduardo C. Garrido-Merchán \textsuperscript{1}, Javier Sánchez-Cañizares\textsuperscript{2}}

\institute{Quantitative Methods Department, Universidad Pontificia Comillas, Madrid, Spain
\email{ecgarrido@icade.comillas.edu} \and
Grupo Mente-Cerebro at ICS, Universidad de Navarra, Navarra, Spain
\email{js.canizares@unav.es}}

\date{July 2022}

\begin{document}

\maketitle

\abstract{
    Integrated information theory (IIT) is a theoretical framework that provides a quantitative measure to estimate when a physical system is conscious, its degree of consciousness, and the complexity of the qualia space that the system is experiencing. Formally, IIT rests on the assumption that if a surrogate physical system can fully embed the phenomenological properties of consciousness, then the system properties must be constrained by the properties of the qualia being experienced. Following this assumption, IIT represents the physical system as a network of interconnected elements that can be thought of as a probabilistic causal graph, $\mathcal{G}$, where each node has an input-output function and all the graph is encoded in a transition probability matrix. Consequently, IIT's quantitative measure of consciousness, $\Phi$, is computed with respect to the transition probability matrix and the present state of the graph. In this paper, we provide a random search algorithm that is able to optimize $\Phi$ in order to investigate, as the number of nodes increases, the structure of the graphs that have higher $\Phi$. We also provide arguments that show the difficulties of applying more complex black-box search algorithms, such as Bayesian optimization or metaheuristics, in this particular problem. Additionally, we suggest specific research lines for these techniques to enhance the search algorithm that guarantees maximal $\Phi$.
}

\section{Introduction}

Together with the Global Neuronal Workspace, Integrated Information Theory (IIT) \cite{Balduzzi2008,Tononi2008,Oizumi2014,tononi2016integrated,Tononi2017} is currently the most prominent approach to obtaining a scientific theory of consciousness, i.e, a theory that can identify the neural correlates of consciousness through some direct or indirect measure. Difficulties in agreeing on experimental setups that may discriminate between different theories have recently led to a sort of adversarial collaboration between them. \cite{Melloni2021} Within said collaborative framework, each theory provides the conditions for it to be experimentally falsified. However, despite such partnership and the obtaining of promising results in clinical tests, \cite{Casali2013,Bodart2017,Haun2017} reaching definitive answers may prove more difficult than initially expected because of the multifactorial complexity of the biological substrate of consciousness, namely the living brain. \cite{Arsiwalla2018a}
\\

The appeal of IIT for computation science and, in general, the AI community relies on its straightforward definition of consciousness as a formal property of any physical system, composed of nodes and interactions, that instantiates it. Consciousness {\it{is}} maximally integrated information $\Phi$; \cite{Tononi2008} the latter quantity being amenable to a strict mathematical definition in a network of interconnected elements that can be thought of as a probabilistic causal graph. Were said statement true, IIT would be providing not only a quantifiable prediction for a system to be conscious but also a heuristic approach to generate conscious systems ---the holy grail of the AI program. Yet it could also happen that IIT offers just a necessary, but not sufficient, condition for the emergence of consciousness in a system. As a logical consequence, a system implementing an artificial intelligence algorithm need not be conscious whatsoever \cite{garrido2022independence} and, hence, need not {\it{understand}} the meaning of the problems that it solves. \cite{garrido2022artificial}
\\

Interestingly, IIT derives its mathematical definition of consciousness from five axioms that fully account for what, phenomenologically, being conscious implies. To wit: The present experience of consciousness exists, is informative, composed, integrated, and excludes other conscious experiences \cite{Oizumi2014}. Said axioms should naturally determine the emergence of a physical scale in which the system, from its own perspective, presents a maximally integrated structure of intrinsic information, in which the information of the whole is maximally irreducible to the sum of information in subsets of any partition of the system. Additionally, the physical state of the substrate instantiating maximal $\Phi$, for the same token, presents a maximal cause-and-effect composition (for the details of the link between measures of integrated information and causality in an interconnected network, one may see \cite{Albantakis2019a,Albantakis2019}).
\\

Whereas IIT continues developing as a research program that strives for finding the best definition of distance between informational contents \cite{Barbosa2021} and including in its framework many other features of consciousness, \cite{Haun2019}
it has been criticized on diverse grounds \cite{Searle,Pautz2019a,Merker2022}. More specifically for our interests in this paper, some authors \cite{Morch2019,Sanchez-Canizares2022} have raised the issue of the so-called 'intrinsicality problem' (IP), namely, the fact that consciousness cannot be an intrinsic property of the system because maximal $\Phi$ crucially depends on the possible existence of bigger values of $\Phi$ if the initial system is appropriately linked to or embedded in larger systems. IP's threat, if true, could manifest that the causality involved in the constitution of the system goes beyond the conceptualization of causality provided by IIT in its current form. \cite{Sanchez-Canizares2022}
\\

Even though the literature on the computation of $\Phi$ in different networks is rapidly increasing and gaining insights on the logical architectures favoring maximal $\Phi$ (basically those consisting of modular, homogeneous, and specialized networks with feedback connections \cite{Oizumi2014}),
the prohibitively large computation times for big networks, scaling with the number of nodes ($n$) as
$\mathcal{O}(n53^n)$,
\cite{mayner2018pyphi} make it necessary the search for alternative heuristics and/or new methods for measuring $\Phi$ in larger systems. \cite{Nilsen2019,Toker2019}
Not much is known about the actual distribution of $\Phi$ over different network types and topologies.
This paper thus explores the benefit of random search algorithms in order to compute $\Phi$. 
If, according to IIT's creators "the quantity and quality of consciousness are what they are," and consequently "the cause–effect structure cannot vary arbitrarily with the chosen measure of intrinsic information" \cite{Barbosa2021}, how  maximal $\Phi$ varies with the number of systems nodes must be a relevant matter to get a glimpse of how consciousness could eventually scale up in networks. Hence, our paper studies the expected increase of maximal $\Phi$ with the number of nodes, paving the way for gaining insights on the network's architectures that favor consciousness and assessing the seriousness of the IP, if IIT holds, in future works.
In such an endeavor, our paper will also show the limits of some optimization techniques for IIT as the number of network nodes increases.
\\

This paper is organized as follows. First, we provide a cursory definition of integrated information, a problem definition of $\Phi$ optimization, and a technical description of the methodology that we have implemented to optimize $\Phi$. Most critically, we provide arguments to justify why our proposed algorithm does not use smarter black-box global optimization techniques such as Bayesian optimization. Then, we show in an experiments section empirical evidence to support the claim that our method is able to optimize $\Phi$. Finally, we present some concluding remarks and refer to possible research lines to optimize $\Phi$ in further work.

\section{Proposed Methodology}
In this section, we begin formally defining integrated information and the problem of optimizing integrated information in the space of transition probability matrices. Then, we show how this problem cannot be solved with common metaheuristics or techniques such as vanilla Bayesian optimization \cite{garrido2021advanced} and introduce a random search algorithm to solve the problem that can be used as a baseline for further research.

\subsection{Integrated information}
Throughout this paper, we use the definitions of integrated information employed in \cite{mayner2018pyphi}. Details regarding the differences between integrated information ($\phi$) for a mechanism ---a set of nodes and interactions in a graph forming a concept--- and conceptual integrated information ($\Phi$) for a constellation of concepts forming a maximally irreducible conceptual structure, can be consulted in \cite{Oizumi2014}. Computations in our paper always refer to  $\Phi$.
\\

We can define integrated information for a mechanism in state $s$ as

\begin{align}
\phi (s) = D [ p(s_{0} \to s) \| p^{MIP} (s_0 \to s)
]
\,,
\end{align}

where $D$ is a measure of the distance between the probability distribution $p$ of obtaining state $s$ as the cause (effect) of future (past) states $s_{0}$ and the correspondent probability distribution $p^{IMP}$ when the system has been optimally partitioned in its weakest links, i.e. decomposed into those parts that are most independent (least integrated), providing a minimum of effectively integrated information for the mechanism as a whole.
\\

When one finds maximal $\phi$ for a mechanism in state $s$ in a system, one has a concept. After determining the structure of concepts in a candidate system, one is allowed to define the conceptual integrated information ($\Phi$) as
$\Phi(C) = D(C \| C^{MIP})$, where $C$ refers to the structure of concepts,  $C^{MIP}$ refers to the optimal partition of the structure of concepts that makes $\Phi(C)$ minimal, and $D$ is an appropriately adapted measure of the distance between the corresponding probability distributions. Different versions of IIT have employed different definitions of the distance between probability distributions ---the Kullback-Leibler divergence for $\phi$ \cite{Balduzzi2008}, the Wasserstein Distance, also known as Earth Mover’s Distance, for both $\phi$ and $\Phi$ \cite{Oizumi2014}, or the Intrinsic Difference \cite{Barbosa2021}. As the PyPhi algorithm favors the Earth Mover’s Distance as the measure of integrated information \cite{mayner2018pyphi}, which quantifies how much two distributions differ by taking into account the distance between system states, we will use it here too.

\subsection{Problem definition}
We now provide a formal definition of the global black-box optimization problem that we need to solve. Concretely, we need to tackle the optimization of a function $\Phi : \mathbb{R}^{2^D,D} \to \mathbb{R}$, from a D-$2^D$-dimensional transition probability matrix, with conditional independencies, space to a real number. 

\begin{align}
\omega^\star = \arg \max_{\omega \in \Omega} \Phi(\omega) \quad s.t. \quad \omega \in \Upsilon \,,
\end{align}

where $\omega^\star$ is the TPM that maximizes $\Phi(\cdot)$, $\omega$ is a TPM of the D-dimensional TPM space, $\Omega \in \mathbb{R}^{2^D,D}$ is the D-dimensional TPM space, $\Upsilon$ is the space of TPM with conditional independencies and $D \in \mathbb{N}^+$ is the number of dimensions of the TPM, whose range in this problem is $\mathbb{N}$. Recall that the objective function $\Phi$ is expensive to evaluate when the number of nodes of the TPM is high. If $\omega \notin \Upsilon$, then $\Phi(\omega) = \emptyset$. Consequently, $\mathbb{R}^{2^D,D}$ is constrained to $\Upsilon$. This is especially problematic, because of the curse of dimensionality. In particular, as $D$ grows, the $\Omega_D$ space will contain higher $\Phi$ TPMs, which is desirable. However, the number of valid TPMs will be lower with respect to all the possible TPMs than in lower dimensional TPM spaces. Consequently, as the evaluation of $\Phi$ is expensive, we have a tradeoff between risking to find high $\Phi$ solutions in a higher dimensional TPM space where the volume of valid TPMs is low or searching in a lower dimensional TPM space where we have a high probability of obtaining a valid TPM but its $\Phi$ is potentially low. More formally, if $\delta(\omega)$ is a delta function whose value is $1$ if $\omega \in \Upsilon$, $D_{min}$ is the lowest dimensionality considered for the number of nodes of the TPM, $D_{max}$ is the highest dimensionality considered for the number of nodes of the TPM and $p(\omega|\Omega_D)$ is the probability of sampling each possible $\omega \in \Omega_D$, we have that:

\begin{align}
 \int \delta(\omega) p(\omega|\Omega_{D_{max}}) d\omega \leq ... \leq \int \delta(\omega) p(\omega|\Omega_{D_{min}}) d\omega \,.
\end{align}

Moreover, as $D$ grows, the cost of evaluating $\Phi$ also grows. Hence, if the budget is constrained by time, and not by iterations, we must be sure about searching in a high dimensional $\Omega$ space. We tackle this issue and define a new algorithm to solve this problem, among others that we further explain, in the following subsection.  

\subsection{Proposed algorithm}
As we have seen in the previous section, the global optimization problem of the $\Phi$ function $\omega^\star = \arg \max_{\omega \in \Omega} \Phi(\omega)$ is not smooth with respect to the smallest variations $\nabla \omega$ that could be performed in the transition probability matrix space $\Omega$. For example, in the case of binary nodes, switching just one node from zero to one can make the $\Phi$ from $\mathbb{R}$ to $\emptyset$. In particular, this is troublesome for some advanced black-box global optimization techniques, as we describe in this section. 
\\

Bayesian optimization \cite{garrido2021advanced,shahriari2015taking} is the state-of-the-art class of methods to solve black-box global optimization problems that iteratively recommend a suggestion $\mathbf{x}$ making a tradeoff between exploration of promising unknown regions of the search space $\mathcal{X}$ and potential good explored solutions based on the predictions made by a probabilistic surrogate model as a Gaussian process \cite{schulz2018tutorial}, which formally is defined as: 

\begin{align}
\mathbf{x}^\star = \arg\min_{\mathbf{x} \in \mathcal{X}} f(\mathbf{x}) \,,
\end{align}

where $\mathbf{x} \in \mathbb{R}^D$ is a candidate point of the hypercube $\mathcal{X} \subseteq \mathbb{R}^D$ and $f(\cdot)$ is the multi-modal function to be optimized. In order to be considered a black-box global optimization problem, $f$ must have no known analytical expression, and hence we are not able to extract gradients; it can be noisy and it is expensive to evaluate. Most critically, if the function $f(\cdot)$ is cheap to evaluate, then it is recommended to use metaheuristics as genetic algorithms \cite{kumar2010genetic}, that also perform an exploration-exploitation tradeoff. The reason is that their execution time to suggest recommendations $\mathbf{x}$ is faster than in the case of Bayesian optimization, whose complexity is a cubic on the number of previous recommendations $\mathcal{O}(N^3)$, where $N$ is the number of previous recommendations.
\\

\SetKwComment{Comment}{/* }{ */}

\begin{algorithm}[htb!]
\caption{Prior guided random search algorithm to optimize $\Phi$ on the D-dimensional TPM space $\Omega$}
\label{alg:phi}
\KwData{$p(\omega), D_{min}, D_{max}, \mu, \epsilon, T$}
\KwResult{$\Phi^\star, \omega^\star$}
$i \gets 0$\;
$\Phi^\star \gets 0$\; 
\While{$i \leq T$}{
$j \gets 0$\;
$\Phi_{list} = []$\;
\While{$j \leq \epsilon$}{ 
    $D_i \gets multinomial\_sample(p(\omega), D_{min}, D_{max})$\;
    $\omega_i \gets uniform\_TPM\_sample(D_i)$\;
    $\Phi^i = \Phi(\omega_i)$\;
  \If{$\Phi^\star \geq \Phi_i$}{
    $\Phi^\star = \Phi_i$\;
    $\omega^\star = \omega_i$\;
  }
  $\Phi_{list} \gets append\_result(\Phi_i, D_i, \Phi_{list})$\;
  }
  $p(\omega) \gets update\_prior(p(\omega), \mu, \Phi_{list})$\;
  }
\end{algorithm}

The problem definition presented in the previous section involves the computation of $\Phi$, which it is very expensive if the number of nodes of the TPM is high. Consequently, this scenario does not present a good fit for metaheuristic algorithms, as they require the evaluation of the objective function to be cheap. Moreover, vanilla Bayesian optimization requires that the function does not vary with respect to its parameters up to a certain $K$-Lipschitz value. \cite{hager1979lipschitz} Critically, in this problem the smallest variation on any elements of the TPM matrix can make $\Phi$ jump from $\mathbb{R}$ to $\emptyset$. Hence this assumption, required by Bayesian optimization, is violated and would need to be specifically modified to tackle this particular problem, whose complexity is even bigger as the space being optimized is $\mathbb{R}^{2^D,D}$, making Bayesian optimization also hard to use. Having reviewed these alternatives, there also exists another reason why those techniques may be inefficient. Concretely, if the $K$-Lipschitz value assumption of Bayesian optimization is violated, the algorithm cannot perform exploitation. Hence, any exploitative behavior on the D-dimensional TPM space $\Omega$ would be inefficient. Consequently, the only alternative that remains is to perform a prior guided random search algorithm, adapted to this particular problem, and encode into the prior our knowledge about the $\Phi$ optimization problem. 
In particular, we enhance a random search procedure instead of a grid search because random search significantly outperforms grid search when the percentage of explained variance on the variable to optimize is not uniform over the explanatory variables \cite{bergstra2012random}. As assuming that the variance of $\Phi$ is uniformly explained by all the nodes for all the possible TPMs is a heavy assumption, we can not rely on uniform grid search, as random search would be a smarter option in this case. Recall that random search outperforms grid search when some variables of the input space explain the objective function better than others. Consequently, we decided to enhance the random search procedure. The prior guided random search illustrated on Algorithm \ref{alg:phi} will obtain $\Phi^\star$, which is an approximation to $\Phi^{best}$, that is, $\Phi^\star \approx \Phi^{best}$, which could only be found using an exhaustive search throughout all the TPM D-dimensional space $\Omega$. Where, in Algorithm $\ref{alg:phi}$, $\mu$ is the adaptation rate of $p(\omega)$, the multinomial a priori distribution of the probability of obtaining $\Phi^\star$ in a certain $D$-dimensional space, $\epsilon$ is the number of iterations that are needed to update $p(\omega)$, $T$ is the budget of total iterations that the user can afford, $D_{min}$ is the lower dimension of the TPM D-dimensional space and $D_{max}$ is the higher dimension of the TPM D-dimensional space. The sampling functions define a probabilistic graphical model where the TPM is modeled as a random variable that depends on the multinomial random variable that encodes the probability distribution assigned to the dimensions of the D-dimensional TPM space. More formally, we obtain $\omega_i$ sampling from the following joint probability distribution:

\begin{align}
\omega = uniform\_matrix(\omega \mid D) \quad multinomial(D \mid \theta, D_{min}, D_{max}) \,.
\end{align}

Sampling from the previous joint distribution of both random variables we can sample TPMs conditioned to the probability of obtaining a TPM belonging to the space of valid matrices $\Upsilon$. The initial state of the TPM necessary to compute $\Phi$ is tested in an iterative fashion until a valid state is found. 
\\

At the end of a batch of $\epsilon$ iterations, we update our beliefs about the $\theta$ parameter values of our multinomial distribution $p(\omega|\theta)$ up to a learning rate value $\mu$. We do so in a Bayesian learning fashion, basically defining a likelihood function $f(\Phi_{list}, \mu)$ that penalizes the dimensions whose results are worse and rewards those whose values are better up to $\mu$. This likelihood function is multiplied by the prior beliefs $p(\omega|\theta)$ and the posterior is obtained by normalizing the previous operation with a normalization constant $\sum_{D_{min}}^{D_{max}} p(\omega|\theta) f(\Phi_{list})$ to ensure that the resulting distribution is a valid multinomial distribution and the prior is conjugate. Finally, we smooth the maximum and the minimum of the vector of probabilities by a quantity $\kappa=0.02$ that could be also modified by the user. The specific parametric likelihood function $f(\Phi_{list}, \mu)$ can be set by the user. Concretely, we just provide as default to linearly penalize or reward each dimension in $0.2 \mu r$ points where $r$ is the ranking of each dimension $D$ that consists of a $[-1,1]$ linear space vector. More formally, if $p(\omega|\theta)_{new}$ is the posterior distribution obtained after the batch of $\epsilon$ iterations, then, to obtain this distribution we perform the following update:

\begin{align}
p(\omega|\theta)_{new} = \frac{f(\Phi_{list}, \mu) p(\omega|\theta)}{\sum_{D_{min}}^{D_{max}} f(\Phi_{list}, \mu) p(\omega|\theta)}
\end{align}

Lastly, we remark that the values of the set of prior beliefs $\theta$, the learning rate $\mu$, and the batch size of iterations $\epsilon$ are hyper-parameters that the user can set and will modify the behavior of the algorithm. Additionally, they also allow the user to set the value of $D_{min}, D_{max}$ to define the TPM search space $\Omega$. In the next section, we illustrate the algorithm performance, being compared with a grid search method and a random search method as baselines.     

\section{Experiments}

In order to estimate the integrated information of a transition probability matrix, we use the PyPhi tool implemented on Python3 \cite{mayner2018pyphi}. In the computation of $\Phi$, to measure distances between probability distributions, we use the Earth Mover's Distance \cite{andoni2008earth}, which is equivalent to the Hamming Distance if the nodes of the graph are binary. We split the experiment section into two parts. First, we show the results obtained by the prior guided random search algorithm with respect to other baselines, just to illustrate how we can obtain graphs whose $\Phi$ score is bigger algorithmically. Driven by the results of these experiments and because of the size of the TPM space that makes it computationally unfeasible to explore, we append another section where we try to statistically make inference about $\Phi$ as the number of nodes grows. We exhaustively search for an initial state for the TPM and retrieve the first feasible state for a randomly sampled TPM, making a point estimation about the mean $\Phi$ of the TPM with the first feasible state found, for computational reasons. All the code, written on Python3, and the experiments are uploaded into the Github repository (https://github.com/EduardoGarrido90/iit\_opt).
\\ 

\subsection{Prior Guided Random Search experiments}

The following experiments model the obtained $\Phi^\star$ by the various methods as a Gaussian random variable with mean $\mu$ and standard deviation $\sigma$. All the experiments are repeated $25$ times to infer their respective values of $\mu$ and standard deviation $\sigma$. Namely, we compare our proposed prior guided random search method with respect to a grid search method and a random search method. The random search method samples uniformly both the biggest dimension of the TPM and the values of the matrix. However, the grid search method uniformly splits the dimensions of the TPM across all the space range, and the binary TPMs are selected by obtaining the binary representations of a linear space of integers from $0$ to $2^n$ where $n$ is the number of nodes which are the columns of the matrices.
\\

\begin{figure}[h]
    \centering
    \includegraphics[width = 0.99\textwidth]{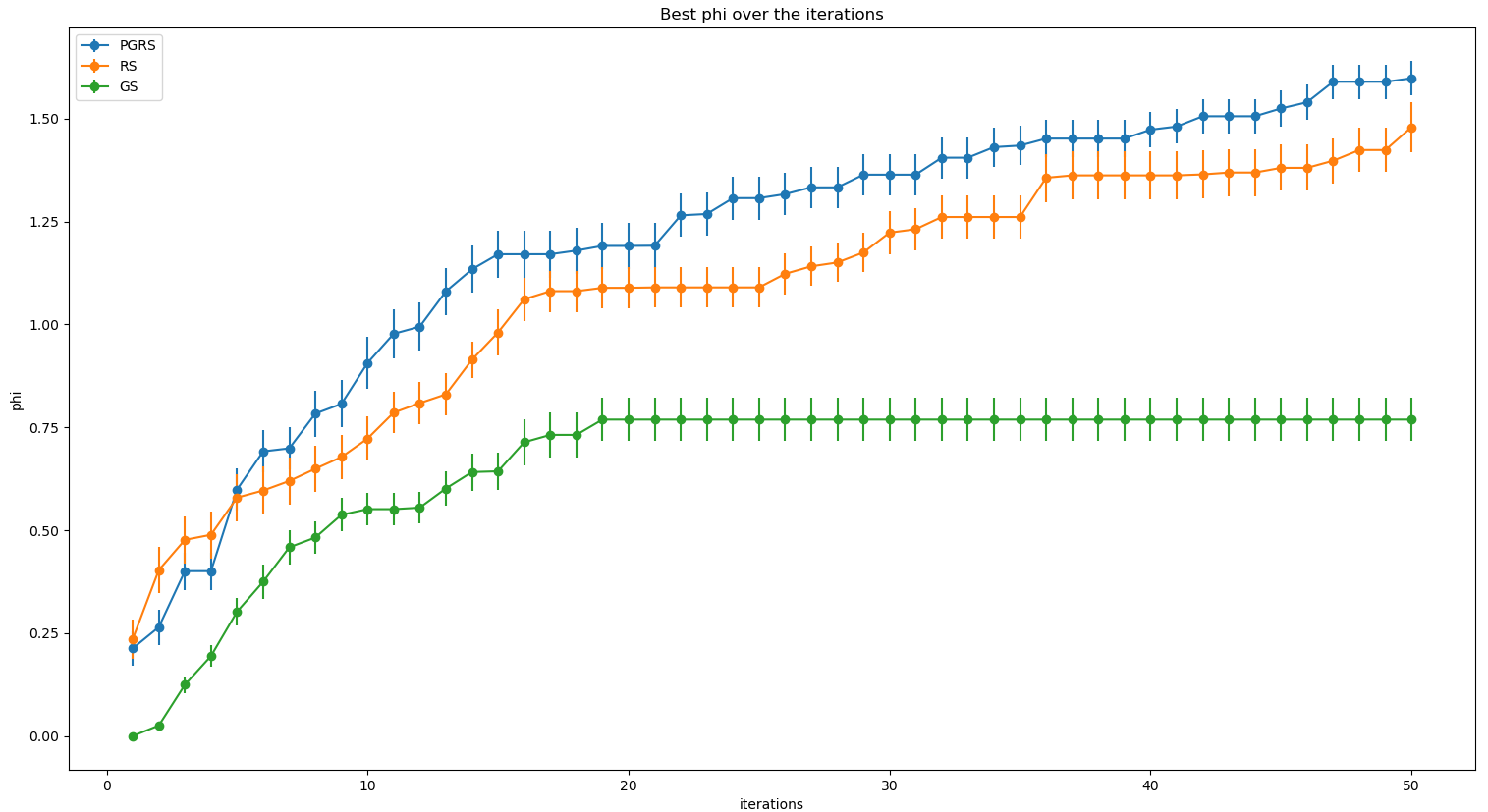}
\caption{Average results of $\Phi$ with the toy experiment configuration of the Prior Guided Random Search method (blue) compared to the Grid (green) and Random search (orange) baselines. Better seen in color.}
    \label{fig:exp1}
\end{figure}

First, we begin with the following experiment, to verify whether all the methods work and provide valid $\Phi$ results. We perform 25 repetitions of the experiment with each method to ensure statistical significance. We set the lower dimension, or the number of nodes, to 3 and the higher dimension to 4. We execute a total of 50 iterations with $\epsilon=5$ for the prior guided search algorithm. We set $\mu=0.1$ and the prior distribution values on the dimensions to $p(\omega) = [0.2, 0.8]$. With these values, we obtain the $\Phi$ results across the iterations shown in Figure \ref{fig:exp1}. The best $\Phi$ found in the search was $3.034082$.
\\

\begin{figure}[h]
    \centering
    \includegraphics[width = 0.99\textwidth]{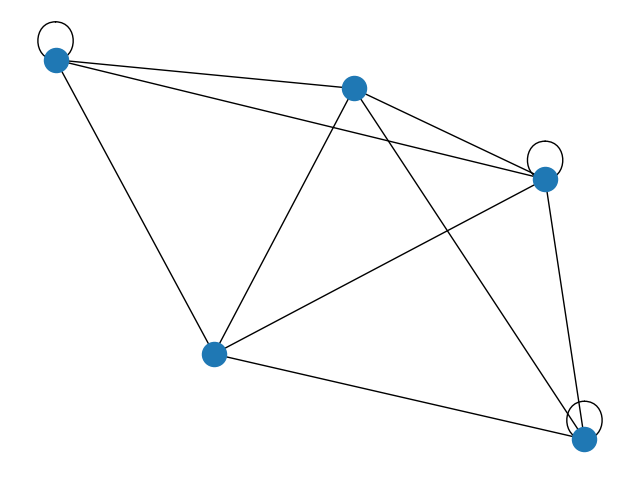}
\caption{Connectivity matrix associated to the best result obtained in the Graphical Experiment.}
    \label{fig:cm}
\end{figure}

\begin{figure}[h]
    \centering
    \includegraphics[height = 0.6\textheight]{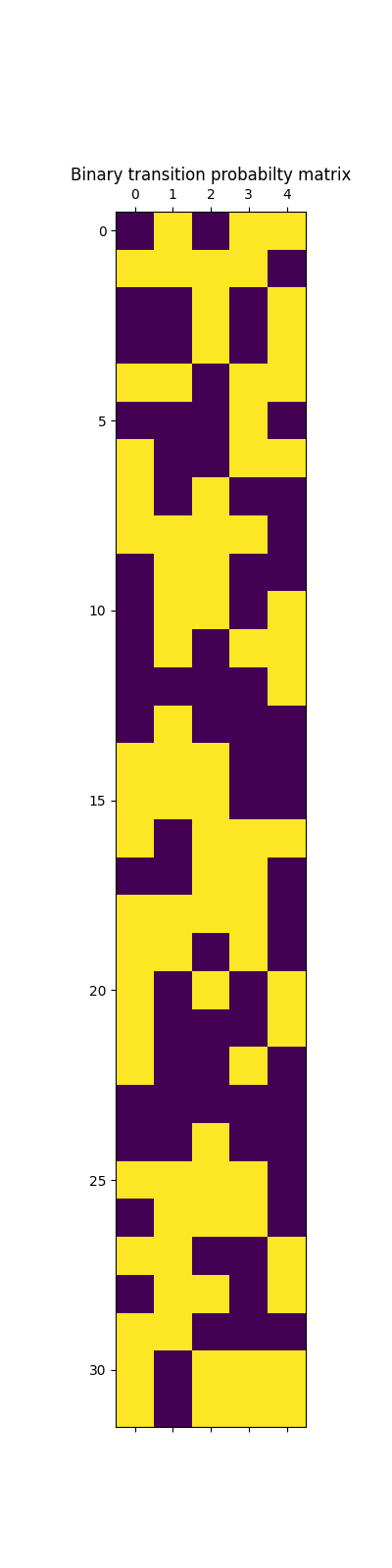}
\caption{Transition probability matrix associated to the best result obtained in the Graphical Experiment. 0 is painted in purple and 1 in yellow.}
    \label{fig:tpm}
\end{figure}

\begin{figure}[h]
    \centering
    \includegraphics[width = 0.99\textwidth]{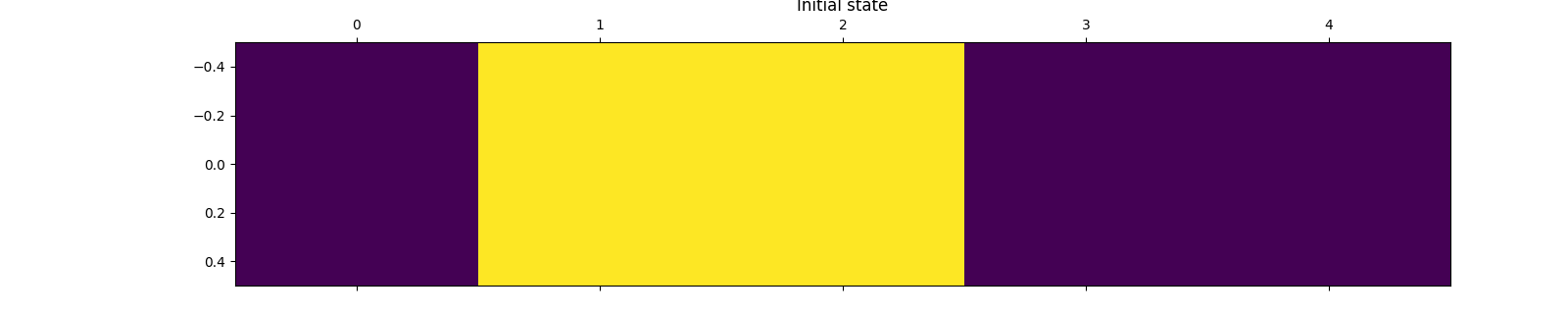}
\caption{Initial state associated with the best result obtained in the Graphical Experiment. 0 is painted in purple and 1 in yellow.}
    \label{fig:state}
\end{figure}

As we can see in Figure \ref{fig:exp1}, the prior guided random search method, plotted in blue, outperform the baselines. The results are statistically significant as the plotted results are the means of $25$ different experiments. As the prior guided random search method performs almost twice as better as the grid search method, a statistical hypothesis testing method will provide a p-value lower than $.01$ to reject the null hypotheses that the differences are random, accepting the alternative hypothesis that both methods have different distributions. Concretely, random search also outperforms grid search. And lastly, the more interesting conclusion is that the prior guided random search method outperforms random search. Regarding grid search, we have found that the restriction of the matrices of the linear grid seems to provide bad results, resulting mostly on invalid TPMs and not being a method that is worth using to optimize $\Phi$ in the $\Omega$ space. Hence, for the next experiments, and also for computational reasons due to the complexity of $\Phi$ with respect to the number of nodes, we will only compare the prior guided random search method with the random search method.
\\
We have performed a little experiment, which we will call Graphical Experiment, of one repetition just to show graphically the obtained transition probability matrix, the connectivity matrix, and the initial state. This experiment involves 100 repetitions with the minimum dimension of nodes being $3$ and the maximum number of nodes $4$. The initial prior is $[0.3, 0.7]$. We obtained, with the prior guided random search algorithm, a graph with $\Phi=2.6823$, with a final prior with value $[0.12, 0.88]$, the connectivity matrix shown on Figure \ref{fig:cm}, the transition probability matrix shown on Figure \ref{fig:tpm} and the state shown on Figure \ref{fig:state}.

\begin{figure}[h]
    \centering
    \includegraphics[width = 0.99\textwidth]{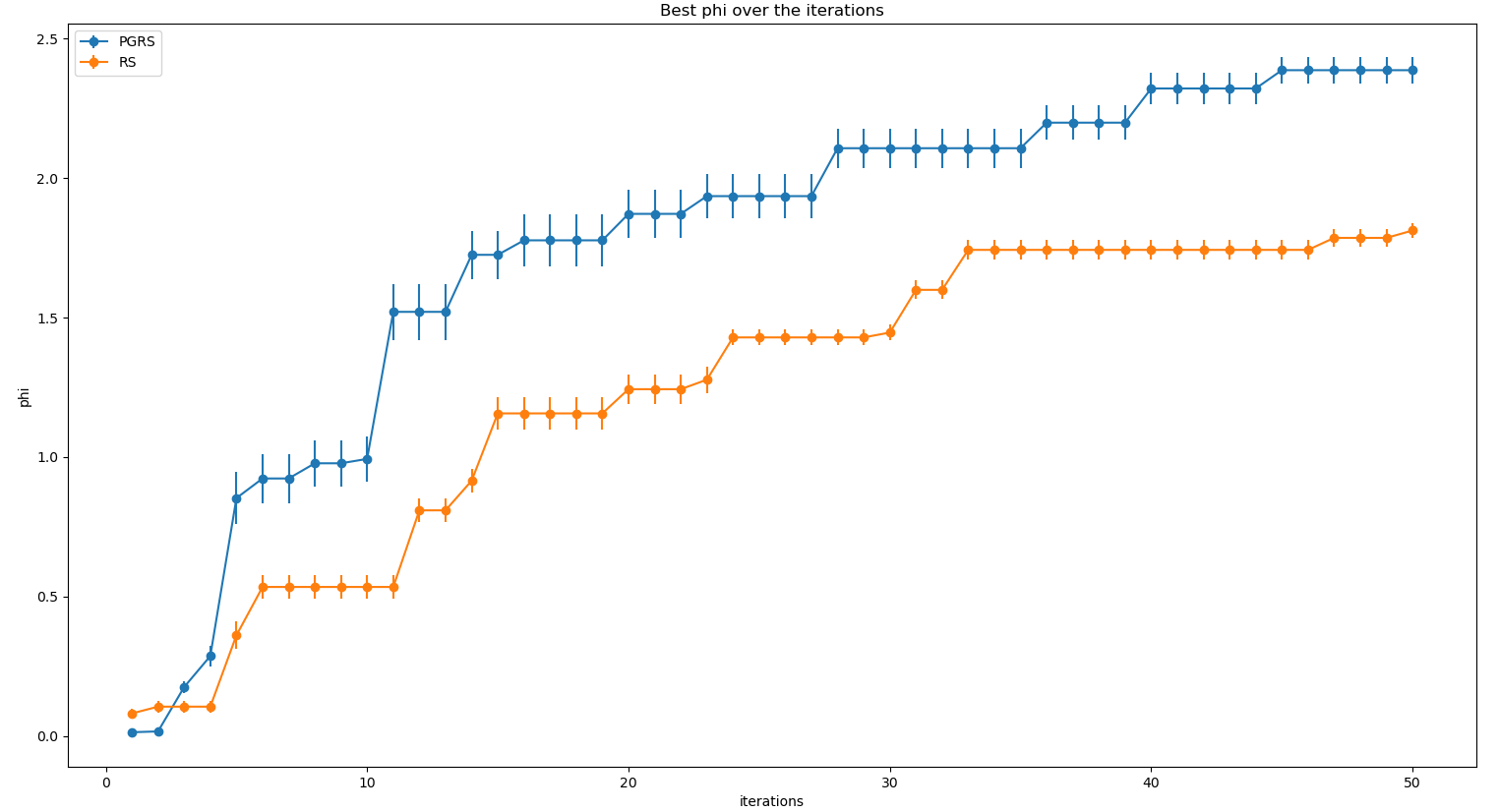}
\caption{Average results of $\Phi$ with the small experiment configuration of the Prior Guided Random Search method (blue) compared to the Random search (orange) baseline. Better seen in color.}
    \label{fig:exp2}
\end{figure}

In the following experiment, we want to verify whether the behavior differences in the method generalize to more than 2 dimensions in the space of TPMs, this time, from 2 nodes to 5. With a prior distribution with weights $p(\omega) = [0.1,0.1,0.3,0.5]$ and only $5$ repetitions to save computational resources and $50$ iterations. We obtain the results plotted in Figure \ref{fig:exp2} where we see how the Prior Guided Random Search algorithm is also outperforming the random search method.
\\

We can see, once again, that both searches are able to find better TPMs according to the $\Phi$ measure once the iterations are executed and that prior guided random search once again outperforms random search, not only on average but in absolute value. In particular, the maximum $\Phi$ found by the prior guided random search method is $3.1681$ and the best result obtained by random search is $2.0832$.
\\

\begin{figure}[h]
    \centering
    \includegraphics[width = 0.99\textwidth]{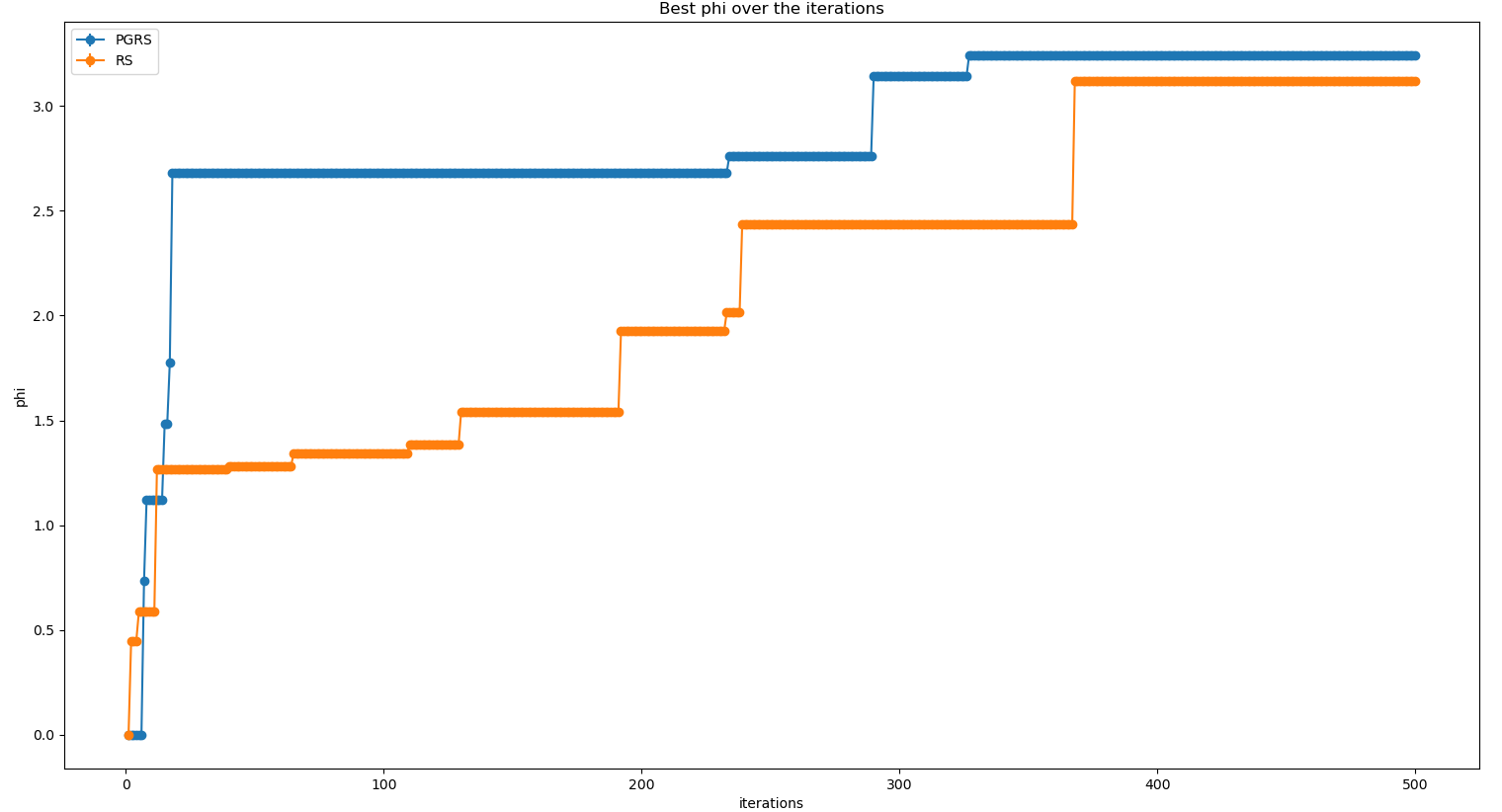}
\caption{Average results of $\Phi$ with the small experiment configuration of the Prior Guided Random Search method (blue) compared to the Random search (orange) baseline. Better seen in color.}
    \label{fig:exp3}
\end{figure}

As we have stated previously, the complexity of evaluating $\Phi$ with respect to the nodes is high, in particular, $\mathcal{O}(n53^n)$, which must be multiplied by $2^n$ in these searches as we look for a feasible state for a randomly sampled TPM. Consequently, and having shown empirical evidence of the usefulness of the random searches method and how our proposed method even outperforms random search for this specific task, our last experiment only involves $1$ repetition of each method, for computational reasons. Being unable to test $\Phi$ in TPMs with $6$ or more nodes due to the complexity of the algorithm, in this experiment, we will check whether better results of $\Phi$ are obtained if the number of nodes and iterations is bigger. The minimum dimension is now $4$ and the maximum dimension is $5$. The prior distribution is $[0.2,0.8]$ and the number of iterations is $500$, with $\mu=1$ and $\epsilon=5$, obtained the results plotted in Figure \ref{fig:exp3}.
\\

We can see how better results are obtained if we have a bigger budget of iterations or time. Moreover, this experiment also shows that the Prior Guided Random Search algorithm outperforms Random search once the prior distribution fits the optimal number of nodes. As the $\Omega$ space is very sparse, a much higher number of iterations would be required to obtain a better approximation to the optimal $\Phi$. More critically, for further work, with more computation power that would enable searches for graphs of more than $6$ nodes, the space would even be more sparse, so much more computation would be needed to find graphs with bigger $\Phi$. However, we have shown with these experiments how our algorithm would adapt to the sparsity of every dimension and keep searching for solutions in the number of nodes that represents the better trade-off between the probability of finding better $\Phi$ and the sparsity of the dimension that makes difficult to find a feasible $\Phi$.

\subsection{Experiments to explore the graph space and make statistical inference of $\Phi$}
The complexity of the space of transition probability matrices makes it computationally intractable to be explored exhaustively. Moreover, in order to measure $\Phi$, we must not only take into account the matrix that encodes the transitions but also the connectivity matrix, that represents the connections, or edges, of the graph. Consequently, the complexity, or size measure, of the space is even higher. 

%Let us formalize this. Let $N$ be the total number of nodes of a graph. The graph has an associated connectivity matrix, that defines its edges, and a transition probability matrix, that defines the next state of the graph, the set of activated nodes, with respect to its previous states. The connectivity matrix has $[N,N]$ shape where each $c_{ij}$ is a dichotomous variable and the transition probability matrix has $[2^N,N]$ shape where we assume that each $t_{ij}$ is also a dichotomous variable. Recall that $t_{ij}$ may also be a probability, but we do not consider those graphs in this paper since they add additional complexity to the problem and restrict the graph space to the binary scenario. If we flatten both graphs, the connectivity matrix space has $2^{N^2}$ possible matrices [¿ESTÁS SEGURO DE ESOS CÁCULOS?]. Each of them may have any possible combination of transition probability matrices [NO ESTOY TAN SEGURO DE ESTO, LA CM RESTRINGE EN ALGUNA MEDIDA LA TPM. PERO SOBRE TODO NO ENTIENDO LA DISCUSIÓN A PARTIR DE LA CM, QUE QUEDA DETERMINADA CON LA TPM], that is, $2^{(N^2*N)} = 2^{N^3}$ possible matrices. Hence, we have $2^{N^2 + N^3}$ possible combinations, which is computationally infeasible to exhaustively explore even for a small number of nodes. For example, for $2$ nodes, there are $4096$ possible graphs, for $3$ nodes, there are $68719476736$ graphs and for $4$ nodes there are $1208925819614629174706176$ combinations. [¿ESTÁS SEGURO DE ESOS NÚMEROS?]

However, we can frame this problem into the classical statistical inference framework. The population is the total of possible combinations for a given number of graphs and we can extract a random sample from it. The goal is to make inferences about the mean $\Phi$ parameter $\mu$ associated with each number of nodes. Then, we can perform statistical hypothesis testing to determine whether the difference of $\Phi$ is statistically significant when $N$ grows. By doing this procedure, we can validate, with the previous set of experiments, that our algorithm is going to work even better when the number of nodes grows, assuming that the same inference will hold when the number of nodes grows, which we can not empirically test since the complexity of computing $\Phi$ when the number of nodes grows is exponential.     

The first toy experiment involves 100 graphs of 3 nodes and 100 graphs of 4 nodes. We obtain an empirical $\Phi$ mean of $0.0517$ in the case of 3 nodes and, in the case of 4 nodes, an empirical mean of $0.1711$. The confidence intervals at 95\% are $[0.0212,0.0822]$ in the case of 3 nodes and $[0.1093,0.2328]$ in the case of 4 nodes. We also perform a 2 sample t-test whose null hypothesis is that both population means are equal and the alternative hypothesis is that they are different. We obtain a t-statistic of $-3.38$ whose associated p-value is $0.0008$. At a confidence level of $\alpha=0.01$ we can reject the null hypothesis that both means are equal, hence accepting the alternative hypothesis that they are different, being the $\Phi$ population mean of graphs of 4 nodes bigger than the $\Phi$ population mean of graphs of 3 nodes. Also, the percentage of infeasible solutions found in the case of $3$ nodes has been $72.6775\%$ and in the case of $4$ nodes has been $84.7094\%$, providing empirical evidence of Eq. 2. To test consistency, we performed the same experiment with a population of $200$ in both groups. In this case, we obtained an even bigger, in absolute terms, a t-test statistic of value $-4.6420$ with associated p-value $4.69e-6$, rejecting the null hypothesis. The obtained empirical means are $0.0569$ and $0.1847$, pretty similar to the ones obtained in the previous experiment with similar confidence intervals.

The second experiment involves a population of 100 graphs of 4 nodes and 100 graphs of 5 nodes. We obtain an empirical $\Phi$ mean of $0.1432$ in the case of 4 nodes and, in the case of 5 nodes, an empirical mean of $0.3267$. The confidence intervals at 95\% are $[0.0822,0.2043]$ in the case of 4 nodes and $[0.2217,0.4317]$ in the case of 5 nodes. We also perform a 2 sample t-test whose null hypothesis is that both population means are equal and the alternative hypothesis is that they are different. We obtain a t-statistic of $-2.94$ whose associated p-value is $0.0036$. At a confidence level of $\alpha=0.01$ we can reject the null hypothesis that both means are equal, hence accepting the alternative hypothesis that they are different, being the $\Phi$ population mean of graphs of 5 nodes bigger than the $\Phi$ population mean of graphs of 3 nodes. Also, the percentage of infeasible solutions found in the case of $4$ nodes has been $84.98\%$ and in the case of $5$ nodes has been a $91.84\%$

\section{Philosophical discussion} 

This work presents an improvement on how to compute $\Phi$ for arbitrary networks and TPMs. Whereas the present calculations still work with toy models with a maximum of 6 nodes, the obtained results with a guided random method strongly suggest that the latter outperforms other methods to explore optimal TPMs as the number of nodes increases. In this sense, it reveals a promising procedure for further progress in the time computation of $\Phi$ and, eventually, for the ascertainment of the validity of IIT as the scientific theory of consciousness in realistic networks. We aim at answering questions like these: Can we quantitatively provide a method to generate consciousness? May this methodology be adapted to deep neural networks if the assumptions of IIT were true, providing deep learning machine consciousness architectures \cite{merchan2020machine}? 
\\

On the other hand, one should keep in mind the intrinsic limitations of the PyPhi algorithm, which obviously affect our results. According to its creators, ``[T]he analysis can only be meaningfully applied to a system that is Markovian and satisfies the conditional independence property. These are reasonable assumptions for the intended use case of the software: analyzing a causal TPM derived using the calculus of perturbations. However, there is no guarantee that these assumptions will be valid in other circumstances, such as TPMs derived from observed time series (e.g., EEG recordings)." \cite{mayner2018pyphi}
\\

The ideas of the last paragraph allow us to dive into additional philosophical problems related to the overall IIT procedures. One of those problems is the origin of networks. In the case of artificial networks (1), the unity of the network is artificially created by network designers, avoiding external influences that can disrupt the workings of the network. In the case of natural networks (2), however, one may wonder who or what causes the unity of the system. Moreover, whereas in case (1) possible consciousness as defined by IIT remains extrinsic, as a bigger maximal $\Phi$ can be reached by adding new nodes to the initial network, it is no longer clear ---indeed very controversial--- that the same procedure works in case (2). Case (2) could be the only scenario in which, in keeping with IIT tenets, consciousness is intrinsic. But if  that is the case, IP still holds for case (2).
\\

Last but not least, our procedure to obtain bigger $\Phi$ as the number of nodes is increased and the system's architecture becomes more complex might not reflect the ontology of case (2). The bottom-line problem is that an optimization procedure might just be an epistemic tool to obtain the magnitude which maximizes an extant, meaningful, quantity. However, the procedure itself need not unravel the ontological processes underlying the reality of said quantity.

\section{Conclusions and further work}

Throughout this paper, we have dealt with integrated information theory as a relatively novel approach to quantifying truly integrated information in a network of interacting nodes. Beyond its technical interest in the field, research in IIT is exploding as it aims to become a scientific theory of consciousness by validating its axioms in artificial and natural networks. Nevertheless, at present, procedures to compute $\Phi$ remain too lengthy. Even if the creation of the PyPhi software has provided IIT researchers with a magnificent tool, computation times increase too quickly with the number of nodes in a network. Yet, studying how maximal $\Phi$ varies when new nodes and interactions are added to the system may prove crucial to see how, allegedly, consciousness scales up and whether the intrinsicality problem presents a true threat to IIT.
\\

Our paper aims to provide a new method to evaluate maximal $\Phi$ with the help of state-of-the-art optimization procedures of network quantities. However, as stated in the technical sections, the requirements of conditional independence for the TPMs charge our procedure with a burden that may not be eased out while keeping all its computational strengths. For the time being, ours remain a toy model that seems promising to pursue in future works.
\\

This work thus opens up several lines of research. To enhance the behavior of the method, we hypothesize that several actions could be taken. First, it may be possible to optimize on a smooth, continuous space of conditional TPMs to use Bayesian optimization or genetic algorithms. This approach would need to formulate an analytical transformation from the space of TPMs into a smooth and continuous space that can be modeled with a Gaussian process, that is the probabilistic surrogate model of Bayesian optimization. By doing so, we can exploit the exploitation-exploration behavior of Bayesian optimization, which may outperform the current prior-guided random search algorithm. If this transformation cannot be eventually found, we can extract the best-observed result by the prior-guided random search algorithm and execute a combinatorial optimization algorithm with the best-observed result as the initial point. This may also provide good results if $\Phi$ varies smoothly as a result of varying the TPM or the CM in a single element. Lastly, as we have seen, $\Phi$ is computationally costly. An interesting line of research would be to find a cheaper approximation of $\Phi$ with respect to a particular graph with a regression graph machine learning algorithm. In order to fit this algorithm it would be necessary to create a huge dataset of graphs annotated by its $\Phi$ values and study whether a regression graph machine learning algorithm is able to accurately approximate $\Phi$.  

\bibliographystyle{acm}

\bibliography{main}

\end{document}